\documentclass[9pt,technote, compsoc]{IEEEtran}

\usepackage{times}
\usepackage{epsfig}
\usepackage{graphicx}
\usepackage[cmex10]{amsmath}
\usepackage{amssymb}
\usepackage{setspace} 
\usepackage{multirow}
\usepackage{xcolor}
\usepackage[normalem]{ulem}
\usepackage{algorithm}
\usepackage{algorithmic}
\usepackage[pagebackref=true,breaklinks=true,letterpaper=true,colorlinks,bookmarks=false]{hyperref}
\usepackage{balance}

\ifCLASSOPTIONcompsoc
  \usepackage[nocompress]{cite}
\else
  \usepackage{cite}
\fi
\ifCLASSINFOpdf
\else
\fi
\newtheorem{thm}{Theorem}

\newtheorem{lem}[thm]{Lemma}

\newcommand{\norm}[1]{\left\lVert#1\right\rVert}

\newcommand\scalemath[2]{\scalebox{#1}{\mbox{\ensuremath{\displaystyle #2}}}}

\newcommand{\colR}[1]{\textcolor{red}{\bf #1}}
\newcommand{\colB}[1]{\textcolor{blue}{\bf #1}}
\newcommand{\tx}[1]{#1}
\newcommand{\ty}[1]{\tiny\bf #1}

\usepackage[pagebackref=true,breaklinks=true,letterpaper=true,colorlinks,bookmarks=false]{hyperref}

\begin{document}

\title{Fast and Robust Fixed-Rank Matrix Recovery}

\author{\small German~Ros*,~Julio~Guerrero,~Angel Sappa,~Daniel Ponsa~and Antonio Lopez%
\IEEEcompsocitemizethanks{\IEEEcompsocthanksitem G. Ros, D. Ponsa and A. Lopez are with the Computer Vision Center \& the Universitat Aut\`onoma de Barcelona, Spain. E-mail: gros@cvc.uab.es.
\IEEEcompsocthanksitem J. Guerrero is with the Department of Applied Mathematics at Universidad de Murcia, Spain.
\IEEEcompsocthanksitem A. Sappa is with the Computer Vision Center, Barcelona, Spain.
\IEEEcompsocthanksitem This work has been supported by the Universitat Aut\`{o}noma de Barcelona, the Fundaci\'{o}n S\'{e}neca 08814PI08, the Spanish government, by the projects FIS201129813C0201; TRA201129454C0301 (eCo-DRIVERS).
}}

\markboth{Fast and Robust Fixed-Rank Matrix Recovery}%
{Ros \MakeLowercase{\textit{et al.}}}
%

\IEEEtitleabstractindextext{%
\begin{abstract}
We address the problem of efficient sparse fixed-rank (S-FR) matrix decomposition, i.e., splitting a corrupted matrix $M$ into an uncorrupted matrix $L$ of rank $r$ and a sparse matrix of outliers $S$. Fixed-rank constraints are usually imposed by the physical restrictions of the system under study. Here we propose a method to perform accurate and very efficient S-FR decomposition that is more suitable for large-scale problems than existing approaches. Our method is a grateful combination of geometrical and algebraical techniques, which avoids the bottleneck caused by the Truncated SVD (TSVD). Instead, a polar factorization is used to exploit the manifold structure of fixed-rank problems as the product of two Stiefel and an SPD manifold, leading to a better convergence and stability. Then, closed-form projectors help to speed up each iteration of the method. We introduce a novel and fast projector for the $\text{SPD}$ manifold and a proof of its validity. Further acceleration is achieved using a Nystrom scheme. Extensive experiments with synthetic and real data in the context of robust photometric stereo and spectral clustering show that our proposals outperform the state of the art.
\end{abstract}

}

\maketitle

\IEEEdisplaynontitleabstractindextext

%
\IEEEpeerreviewmaketitle


\section{Introduction}
\label{intro}

\IEEEPARstart{S}{ystems} with fixed-rank constraints exist in many applications within the fields of computer vision, machine learning and signal processing. Some examples are: photometric stereo, where depth is estimated from a still camera that acquires images of an object under different illumination conditions, leading to a rank constraint; motion estimation, where the type of motion of the objects defines a rank.

This paper addresses the problem of efficient sparse fixed-rank (\textit{S-FR}) matrix decomposition, i.e.: given a matrix $M$ affected by outliers, this is, gross noise of unknown magnitude, we aim to recover an uncorrupted matrix $L$ and a sparse matrix $S$ such that  $M = L + S$ and $\text{rank}(L) = r$, with $r$ known beforehand, as defined in  (\ref{eq:fr}),

\begin{equation}
\label{eq:fr}
\text{min}_{L, S} \norm{S}_{\ell_1} \text{s.t. } M = L + S,\  \text{rank}(L) = r.\\
\end{equation}

\textit{S-FR} matrix recovery is intimately related to the sparse low-rank (\textit{S-LR}) recovery problem (\ref{eq:rpca}), for which algorithms such as Robust Principal Component Analysis (RPCA)~\cite{Candes11} and Principal Component Pursuit (PCP)~\cite{PCP} are well known due to their extraordinary capabilities to solve it and their application to a wide range of problems.

\vspace{-2mm}
\begin{equation}
\label{eq:rpca}
\text{min}_{L, S} \norm{L}_* +  \lambda \norm{S}_{\ell_1} \text{s.t. } M = L + S.\\
\end{equation}
\vspace{-3mm}

Robust \textit{S-FR} recovery might seem a simpler case of \textit{S-LR} decomposition, or even a straightforward derivation. However, \textit{S-FR} recovery is a hard problem that involves a highly non-convex constraint due to the rank imposition. This factor is not present in the \textit{S-LR} decomposition problem due to the nuclear norm relaxation. Therefore, a careful design is needed in order to produce a stable \textit{S-FR} decomposition method with a good convergence rate.

In addition to the convergence speed, achieving efficient and scalable \textit{S-FR} decompositions requires algorithms with very low computational complexity per iteration. The main bottleneck of these algorithms is the enforcement of the correct rank or its minimization, a step that usually requires the use of a TSVD or an SVD per iteration, which complexity is $\mathcal{O}(mnr)$ for a $m \times n$ matrix of rank $r$. How to reduce this bottleneck is a line of research that has been recently targeted by several works such as \cite{Halko}\cite{LMAFIT}\cite{ROSL}, showing interesting ideas leading to algorithms with quadratic and linear complexities with respect to the input matrix size. The key lessons to learn from these works are two: \textbf{(i)} the factorization of large-scale problems into products of small size matrices~\cite{AS}; and \textbf{(ii)} the use of a sub-sampled version of the input matrix to produce fast and accurate approximations of the solution~\cite{ROSL}. 

Our work has been influenced by these concepts and several ideas drawn from state-of-the-art differential geometry techniques. We have experimented with the mentioned concepts and improved upon them in order to create an efficient and precise \textit{S-FR} decomposition algorithm suitable for large scale problems. In this respect we present the following contributions: (\textbf{i}) an optimization method, named FR-ADM\footnote{Code is available at {\small\url{https://github.com/germanRos/FRADM}}} (Fixed-Rank Alternating Direction Method), that solves \textit{S-FR} problems following an ADM scheme to minimize an Augmented Lagrangian cost function; (\textbf{ii}) a novel procedure to impose fixed-rank constraints through a very efficient polar factorization, named \textit{FixedRankOptStep}, which is superior in convergence, stability and speed than the bilinear counterparts used by state-of-the-art methods and; (\textbf{iii}) the use of a simple projector to impose SPD constraints efficiently along with a novel proof of its validity.

We show that our method, based on the \textit{FixedRankOptStep} procedure, outperforms in time, accuracy and region of applicability current state-of-the-art methods. Furthermore, we also show that our proposal FR-ADM can benefit from Nystrom's method~\cite{Nystrom1} to improve its computational efficiency while maintaining a good level of accuracy. These results are supported by thorough experimentation in synthetic and real cases, as detailed in Sec.~\ref{sec:results}.

\section{Summary of Notation}
\label{sec:notation}

Capital letters, such as $M$ represent matrices, while vectors are written in lower-case. $M^T$ stands for the matrix transpose, $M^+$ for its pseudo-inverse and $\textit{tr}(M)$ is the matrix trace operator. $\sigma_k$ stands for the $k$-th largest singular value of a given matrix. The indexation of the $i$-th row and the $j$-th column is defined as $M_{ij}$. Matrix sub-blocks of $M$ are referred to as $M_{[r_1:r_m, c_1:c_n]}$ to index from row $r_1$ to $r_m$ and column $c_1$ to $c_n$. $\norm{M}_F = \sqrt{\textit{tr}(M^T M)}$ is the Frobenius norm and $\norm{M}_{\ell_1} = \sum_{ij} |M_{ij}|, \norm{M}_* = \sum_i \sigma_i$ are the entry-wise $\ell_1$-norm and the matrix nuclear norm, respectively. $I_{m}$ and $I_{m\times n}$ are the square and the rectangular identity matrices. $\text{St}_{m,r}$, is the Stiefel manifold of matrices $U \in \mathbb{R}^{m \times r}$ with $U^T U = I_r$. $\text{SPD}_r$ and $\text{SPSD}_r$ stand for the $r \times r$ Symmetric (Semi-)Positive Definite matrices, respectively. $\mathcal{F}^{(r)}_{m,n}$ is the fixed-rank manifold of matrices $L \in \mathbb{R}^{m \times n}$ with $\text{rank}(L)=r$ and $\mathbb{R}_*^{m \times r}$ is the set of full-rank matrices. $O_r$ stands for the Orthogonal group, but be careful, since $\cal{O}$ is also used to describe the complexity of algorithms in big-O notation. We also make use of some proximity operators and projectors defined as: $\textit{Sym}(M) = \frac{1}{2}(M+M^T)$, the symmetric part of M. $P_{ST}[M] = \text{max}(0, M-\delta) + \text{min}(0, M+\delta)$ for the standard  soft-thresholding (promotes sparsity); $P_{O}[\cdot]$ for the projector onto the Stiefel manifold, and $P_{\text{SPD}}[\cdot]$ for the projector onto the $\text{SPD}$ manifold (these are defined in Sec.~\ref{sec:polar}).

\section{Related Work}
\label{sec:related}

The Accelerated Proximal Gradient (APG)~\cite{APG} will serve as our starting point within the plethora of methods present in the literature. This method, although it is not the first one proposed to solve the RPCA problem (see for instance FISTA~\cite{FISTA}), is an appealing combination of concepts. It approximates the gradients of a given cost function to simplify its optimization and improves its convergence. It also includes Nesterov updates~\cite{Nesterov1} and the critical continuation scheme~\cite{APG}, which all together lead to a method with sub-linear convergence $\mathcal{O}(1/k^2)$, where $k$ is the number of iterations. Its computational complexity per iteration is $\mathcal{O}(n^3)$ for $n\times n$ matrices. Afterwards, authors of \cite{ALM} proposed the Augmented Lagrangian Multiplier method (ALM) in two flavours. First they present the exact ALM (eALM), which uses an Alternating Direction Method (ADM) to minimize an Augmented Lagrangian function in a traditional and exact way. Then, an inexact version is also proposed (iALM), which approximates the original algorithm to reduce the number of times the SVD is used. The convergence rate of eALM depends on the update of $\mu_k$, the penalty parameter of the Augmented Lagrangian. When the sequence $\{\mu_k\}_{k=1}^{k_\text{max}}$ grows geometrically following the continuation principle, eALM is proven to converge Q-linearly $\mathcal{O}(1/\mu_k)$. For iALM, there is not proof of convergence, but it is supposed to be Q-linear too. Both methods have computational complexity of $\mathcal{O}(n^3)$ per iteration.

Recently, ALM was extended in~\cite{LADM}, which included a factorization technique along with a TSVD from the PROPACK suite~\cite{PROPACK} to achieve a complexity of $\mathcal{O}(rn^2)$ per iteration. The bottleneck caused by the TSVD has also been addressed via random projections, leading to the efficient Randomized-TSVD (R-TSVD)~\cite{Halko}. However, although being more efficient than the regular TSVD, results are considerably less accurate. The idea of including a factorization of the data was then improved by LMAFIT~\cite{LMAFIT}, which uses a bi-linear factorization to produce two skinny matrices $U_{m \times k}$ and $V_{k \times n}$, such that $L=UV$, to speed up the process. A similar concept was used in the Active Subspace method (AS)~\cite{AS}, but in this case the bi-linear factorization is given by $ Q_{m \times k}$ and $J_{k \times n}$, such that $Q \in St_{m,k}$. This formulation turns out to be very useful when $m \gg n \gg k$, leading to a complexity per iteration of $\mathcal{O}(mnk)$. Unfortunately, $k$ is an upper bound for the actual rank of $L$ and needs to be given by the user. This is not suitable for \textit{S-LR} scenarios, but fits perfectly in the \textit{S-FR} framework. Another point to highlight about LMAFIT and AS is the utilization of closed-form projectors to impose constraints like orthogonality, low-rank and sparsity. This algebraical way of optimizing functions differs from the geometrical counterparts in the literature on manifold optimization (see~\cite{optiManifold}\cite{Meyer}). Substituting all the required machinery to perform differential geometry (e.g., retractions, lift maps, etc.) by projectors seems a good idea from the point of view of efficiency. However, this method is not absent of problems. The factorization in AS is highly non-convex, an issue that influences the number of iterations required for the convergence of the method, which is notably higher than eALM, despite having the same theoretical convergence rate $\mathcal{O}(1/\mu_k)$.


One of the contributions of our work is to improve the convergence of fixed-rank projection methods. To this end we employ a polar decomposition as in~\cite{Meyer}. This polar decomposition offers us the possibility of exploiting the manifold structure of fixed-rank problems as the product of two Stiefel and an SPD manifold. $\mathcal{F}^{(r)}_{m,n} = (\text{St} \times \text{SPD} \times \text{St})/O_r$. However, we deviate from~\cite{Meyer} to propose more efficient expressions that make use of projectors to speed up the process, giving rise to a better convergence. We also consider worth highlighting a key tool described in the recent work~\cite{ROSL}. There, the authors follow a strategy that resembles the one described in~\cite{AS}, but they add a sub-sampling step based on the Nystrom's method~\cite{Nystrom1} that leads to a linear complexity $\mathcal{O}(r^2(m+n))$ per iteration. We borrow this idea to further speed up our optimization.
\vspace{-1mm}
\section{Sparse Fixed-Rank Decomposition}
\label{sec:approach}

We propose the resolution of the non-convex program in (\ref{eq:decomp}) as a direct way to perform the sparse and fixed-rank decomposition---note that (\ref{eq:decomp}) is equivalent to the program (\ref{eq:fr}) defined in Sec.\ref{intro}. 

\begin{equation}
\label{eq:decomp}
\text{min}_{L, S} \norm{S}_{\ell_1}, s.t.\  M = L + S,\  L \in \mathcal{F}^{(r)}_{m,n},
\end{equation}

The optimization of (\ref{eq:decomp}) is carried out over an Augmented Lagrangian function, leading to (\ref{eq:lagrange}). $Y$ stands for the Lagrange multiplier and $\mathcal{F}^{(r)}_{m,n}$ represents the fixed-rank manifold of rank $r$.

\vspace{-3mm}
\begin{equation}
\scalemath{1}{
\mathcal{L}(L, S, Y, \mu) = \frac{\mu}{2}\norm{M-L-S}_F^2 + \norm{S}_{\ell_1} + \langle Y, M-L-S \rangle \nonumber
}\end{equation}
\vspace{-4mm}
\begin{equation}
\label{eq:lagrange}
s.t.\  L \in \mathcal{F}^{(r)}_{m,n}
\end{equation}

To efficiently solve (\ref{eq:lagrange}) we utilize an ADM scheme~\cite{ALM} endowed with a continuation step, as presented in Algorithm~\ref{ADM}. The update of the fixed-rank matrix $L$ is obtained via the \textit{FixedRankOptStep} algorithm that implements the proposed polar factorization. For the sparse matrix $S$, the standard soft-thresholding is used. Notice that $\mu_k$ is updated following a geometric series (Alg.\hyperref[ADM]{\ref{ADM}.\#9}) in order to achieve a Q-linear converge rate $\mathcal{O}(1/\mu_k)$~\cite{ALM}. Despite having the same asymptotic convergence order as LMAFIT, AS and ROSL, our method FR-ADM takes less iterations to converge, due to the accuracy of the novel \textit{FixedRankOptStep}. This is especially important for the cases where the magnitude of the entries of $S$ are similar to those of $L$, a  challenging situation that other state-of-the-art methods fail to address correctly. We provide empirical validation for this claim in Sec.~\ref{sec:results}.

\vspace{2mm}
\setlength{\intextsep}{4pt}
\setlength{\textfloatsep}{10pt}
\begin{algorithm}
 \algsetup{linenosize=\small}
  \small
\begin{algorithmic}[1]
        \REQUIRE Data matrix $M\in \mathbb{R}^{m\times n}$, $r$ (rank)
    \STATE $k \leftarrow 1, S_k \leftarrow 0_{m \times n}, L_k \leftarrow 0_{m \times n}, Y_k \leftarrow 0_{m \times n}$
    \STATE $\mu_k \leftarrow 1, \rho > 1, \bar{\mu} \leftarrow 10^9, U_0 \leftarrow I_{m \times r}, B_0 \leftarrow I_{r}, V_0 \leftarrow I_{r \times n}$
    \WHILE{not converged}
	\STATE $L_{k+1} \leftarrow \textrm{arg min}_{L\in \mathcal{F}_r} \mathcal{L}(L, S_k, Y_k, \mu_k)$
	\STATE \ \ \  \ \  \  \  \  \ \  \ $= \textit{FixedRankOptStep}(M - S_k + \frac{1}{\mu_k} Y_k, U_k, B_k, V_k)$
	\STATE $S_{k+1} \leftarrow \textrm{arg min}_{S\in \mathbb{R}^{m \times n}} \mathcal{L}(L_{k+1}, S, Y_k, \mu_k)$
	\STATE \ \ \  \ \  \  \  \  \ \  \ $= P_{\text{ST}_{1/\mu_k}}(M - L_{k+1} +  \frac{1}{\mu_k} Y_k)$
	\STATE $Y_{k+1} \leftarrow Y_k + \mu_k (M - L_{k+1} - S_{k+1})$
	\STATE $\mu_{k+1} \leftarrow \text{min}(\bar{\mu}, \rho \mu_k)$
	\STATE $k \leftarrow k + 1$
	\ENDWHILE
	\RETURN{$L_k$, $S_k$.}
\end{algorithmic}
\caption{\textit{FR-ADM}}
\label{ADM}
\end{algorithm}

An adapted version of FR-ADM, referred as FR-Nys, is also provided. FR-Nys exploits Nystrom's subsampling method (\!\!\cite{Nystrom1}\cite{ROSL}), to further speed up computations. This method is presented in Algorithm~\ref{ADM-Nys} and follows the recipe given in~\cite{ROSL}. 

\setlength{\floatsep}{0pt}
\setlength{\textfloatsep}{5mm}
\setlength{\intextsep}{6pt}
\begin{algorithm}
 \algsetup{linenosize=\small}
  \small
\begin{algorithmic}[1]
        \REQUIRE Data matrix $M\in \mathbb{R}^{m\times n}$, $r$ (rank)
    \STATE $\tilde{M} \leftarrow \textit{random-row-shuffle}(M)$
    \STATE $(L_L, S_L) \leftarrow \text{FR-ADM}(\tilde{M}_{[1:m, 1:l]},\  r)$,\ \  for $l = k r$
    \STATE $(L_T, S_T) \leftarrow \text{FR-ADM}(\tilde{M}_{[1:l, 1:n]},\  r)$,\ \ \  for $l = k r$
    \STATE $L \leftarrow L_L \ {L_L}^+_{[1:l, 1:l]}\  L_T$,\ \   $S \leftarrow \tilde{M} - L$
	\RETURN{$L$, $S$.}
\end{algorithmic}
\caption{\textit{FR-Nys}}
\label{ADM-Nys}
\end{algorithm}

Nystrom's scheme proceeds by randomly shuffling the rows of $M$ producing $\tilde{M}$. Then, the top and the left blocks of $\tilde{M}$ are processed separately by using FR-ADM (Alg.\ref{ADM}). Notice that these blocks are chosen of size $m \times l$ and $l \times n$, where $l$ has to be a number larger than the expected matrix rank. In our case $k = 10$. Finally the independently recovered matrices $L_L$ and $L_T$ are combined to produce $L$ and $S$.

\subsection{Polar Factorization on the Fixed-Rank Manifold}
\label{sec:polar}

Imposing rank constraints requires an efficient way of computing the projection of an arbitrary matrix  $M \in \mathbb{R}^{m \times n}$ with arbitrary rank $k\geq r$ 
onto the fixed-rank manifold $\mathcal{F}^{(r)}_{m,n}$. A simple solution is provided by the EckartÃ¢â‚-¬â€œYoung theorem~\cite{SVD}, which shows that the optimization problem (\ref{eq:Eck}):

\begin{equation}
\textrm{min}_{\textrm{rank}(L)= r}\norm{M-L}_F^2,
\label{eq:Eck}
\end{equation}
\vspace{-2mm}

\noindent is solved by the truncated SVD (TSVD) of $M$. Despite the success of the TSVD as a tool for producing low-rank approximations, and the many available improvements, as for instance the usage of random projections\cite{Halko}, some problems require the computation of many TSVDs (typically one per iteration) of very large matrices. Thus an efficient alternative to the usual TSVD algorithm is required. 

In this section we propose the method \textit{FixedRankOptStep} (Algorithm~\ref{Alg-FRStep}), which computes a fast approximate solution to the projection of a matrix onto the fixed-rank manifold, like the given by the TSVD but much faster.  
Additionally, in the Appendix we also propose the method \textit{FixedRankOptFull} (Algorithm~\ref{Alg-FROpt}) that can be seen as a series of iterations of the \textit{FixedRankOptStep} algorithm, providing a solution with a prescribed accuracy and with Q-linear convergence rate to the minimization problem (\ref{eq:Eck}) living on $\mathcal{F}^{(r)}_{m,n}$. 
The \textit{FixedRankOptStep} algorithm is suitable for large-scale problems where many TSVDs of large matrices are required, and an approximate solution is faster and enough for convergence, as we will show later.


Following~\cite{Meyer}, we use a polar factorization on $\mathcal{F}^{(r)}_{m,n}$ suggested by the TSVD. Given a matrix 
$L \in \mathbb{R}^{m \times n}$ of rank $r$, its TSVD factorization is

\vspace{-2mm}
\begin{equation}
 L=U\Sigma V^T,
\end{equation}

\noindent where $U\in \textrm{St}_{m,r}$, $V\in \textrm{St}_{n,r}$ and $\Sigma=\text{diag}(\sigma_1,\ldots,\sigma_r)$. Then, a transformation 

\begin{equation}
(U,\Sigma, V)\rightarrow (U O,O^T\Sigma O, V O),
\end{equation}

\noindent where $O\!\in\! O_r$, does not change $L$, and allows to write it as 

\begin{equation}
L=U'B V'^T,
\end{equation}

\noindent where now $B=O^T\Sigma O\in \textrm{SPD}_r$, $U'=UO$, and $V'=VO$. Thus, the fixed-rank manifold can be seen as the quotient manifold 
$(\textrm{St}_{m,r}\times \textrm{SPD}_r\times \textrm{St}_{n,r})/O_r$. From this, we reformulate (\ref{eq:Eck})  in $\mathcal{F}^{(r)}_{m,n}$ as the solution of
 (\ref{eq:MinimizationFixedRank}).

\setlength{\textfloatsep}{0mm}
\vspace{-1mm} 
\begin{equation}
\textrm{min}_{U\in \textrm{St}_{m,r},B\in \textrm{SPD}_r,V\in \textrm{St}_{n,r}}\norm{M-UB V^T}_F^2.
\label{eq:MinimizationFixedRank}
\vspace{1mm}
\end{equation}

The \textit{FixedRankOptStep} algorithm performs a single step of an alternating directions minimization (ADM)
on each of the submanifolds $\textrm{St}_{m,r}$, $\textrm{St}_{n,r}$ and $\textrm{SPD}_r$ (Algorithm~\ref{Alg-FRStep}).

\setlength{\textfloatsep}{0pt}
\begin{algorithm}
 \algsetup{linenosize=\small}
  \small
\begin{algorithmic}[1]
        \REQUIRE Data matrix $M\in \mathbb{R}^{m\times n}$, previous values $U_0\in \textrm{St}_{m,r}$, $B_0\in \textrm{SPD}_r$, $V_0\in \textrm{St}_{n,r}$
	\STATE $\bar{U} \leftarrow \textrm{arg min}_{U\in \textrm{St}_{m,r}}\norm{M-UB_0 V_0^T}_F^2 = P_\textrm{O}[MV_0B_0]$
	\STATE $\bar{V} \leftarrow \textrm{arg min}_{V\in \textrm{St}_{n,r}}\norm{M-\bar{U}B_0 V^T}_F^2 = P_\textrm{O}[M^T\bar{U}B_0]$
	\STATE $\bar{B} \leftarrow \textrm{arg min}_{B\in \textrm{SPD}_r}\norm{M-\bar{U}B \bar{V}^T}_F^2=Sym(\bar{U}^TM\bar{V})$	
	\RETURN{$\bar{U}\in \textrm{St}_{m,r}$, $\bar{B}\in \textrm{SPD}_r$, $\bar{V}\in \textrm{St}_{n,r}$. }
\end{algorithmic}
\caption{\textit{FixedRankOptStep} Algorithm}
\label{Alg-FRStep}
\end{algorithm}

\subsubsection{\bf Minimization on the Stiefel Manifold}

The minimization subproblems on Stiefel manifolds involving $U$ and $V$ in Algorithm~\ref{Alg-FRStep},  are not the standard  Stiefel Procrustes Problem~\cite{Stiefel-Procrustes}. Here, the Stiefel matrix is left-multiplying instead of right-multiplying, as usually, which allows to provide a fast closed-form solution by using the Orthogonal Procrustes Problem (OPP)~\cite{Orth-Procrustes}, as shown in (\ref{eq:OrthogonalProcrustes}):

\setlength{\textfloatsep}{-8pt}
\begin{equation}
\textrm{min}_{U\in \textrm{St}_{m,r}}\norm{M-UBV^T}_F^2 \Rightarrow\quad U=P_{\textrm{O}}[MVB],
\label{eq:OrthogonalProcrustes}
\end{equation}

\noindent where $P_{\textrm{O}}[A]$ denotes the projector onto the Stiefel Manifold. This can be efficiently computed through a skinny SVD as $A= Q\Sigma S^T \,,Q\in\textrm{St}_{m,r}\,,S\in O_r \Rightarrow  P_{\textrm{O}}[A]=QS^T$. Alternatively, if $\text{rank}(A)=r$ (maximal rank, as we shall assume in the following), it can be computed as $P_{\textrm{O}}[A]=A(A^TA)^{-1/2}$. This shows that $P_{\textrm{O}}[A]$ always exists and it is unique. A similar result holds for the minimization of $V$ by simply transposing (\ref{eq:OrthogonalProcrustes}).

\subsubsection{\bf Minimization on the SPD manifold} 

The minimization subproblem on the SPD manifold is more challenging. The reason is that, although convex, the SPD manifold is an open manifold and therefore the existence of a global minimum is not guaranteed. Its closure is the SPSD manifold, and there the existence of a solution is neither guaranteed. However, we shall see that in our case there exists a minimun in $\text{SPD}_r$. Let us analyse this by first introducing a novel projector onto the SPD manifold. To this end we consider the SPD Procrustes Problem~\cite{SPDProcrustes} (\ref{SPDProcrustes}):

\vspace{-3mm}
\begin{equation}
\scalemath{0.98}{
\textrm{min}_{B\in \textrm{SPD}_r}\norm{M-UBV^T}_F^2 \Rightarrow  B=P_{\textrm{SPD}}[U^TMV],}
\label{SPDProcrustes}
\vspace{-0mm}
\end{equation}

\noindent where the projector $P_{\textrm{SPD}}[A]$ is simply given by $P_{\textrm{SPD}}[A]=\textit{Sym}(A)$. In general, the solution of the SPD Procrustes Problem requires solving a Lyapunov equation~\cite{SPDProcrustes}, but in our case is simpler since $U$ and $V$ are orthogonal. Although in general there is not guarantee that $B=\textit{Sym}(U^TMV)$ is positive definite, we can assure it for our formulation, see the Appendix.

\vspace{-1mm}
\subsection{Convergence Analysis of FR-ADM}

Since the optimization problem (\ref{eq:decomp}) is highly non-convex, a global convergence theorem as in eALM \cite{ALM} cannot be given. However, a weak covergence result similar to iALM or that of LMAFIT~\cite{LMAFIT} (where there is no nuclear norm minimization) can be given. For that purpose, let us state the first-order optimality conditions for the 
constrained minimization problem (\ref{eq:decomp}):

\vspace{-3mm}
\begin{eqnarray}
\label{KKT}
U^TY &=& 0 \nonumber\\
YV &=& 0 \nonumber\\
S &=& P_{ST_{1/\mu}}(S+Y/\mu)\\
M &=& L + S \nonumber
\end{eqnarray}

\noindent where $L=UBV^T$, $\mu>0$ and $Y$ is a Lagrange multiplier. Then we can prove that:

\vspace{3mm}
\begin{thm}
 If the sequence of iterates 
generated by Algorithm~\ref{ADM} converges to a point $(U^*,B^*,V^*,S^*,Y^*)$, this point satisfies the conditions 
(\ref{KKT}) and therefore is a local minimum of (\ref{eq:decomp}).
\end{thm}

\noindent \textit{Proof}: Using Algorithm~\ref{ADM}, and given a projector $P_Q \equiv Q Q^T$, we have:

\begin{equation}
\scalemath{1}{
\begin{split}
& Y_{k+1}-Y_k \rightarrow 0\quad \Rightarrow\quad L^*+S^*= M \nonumber \\
& S_{k+1}-S_k\rightarrow 0 \quad \Rightarrow\quad S^*=P_{ST_{1/\mu}}(S^*+{Y^*}/{\mu})\\
&(P_{U_{k+1}}-P_{U_k})(M-S_k+Y_k/\mu)V_k\rightarrow 0 \Rightarrow P^\perp_{U^*}Y^*V^* = 0\\
&(P_{V_{k+1}}-P_{V_k})(M^T-S^T_k+Y^T_k /\mu)U_k\rightarrow 0 \Rightarrow P^\perp_{V^*}{Y^*}^TU^* = 0\\
&B_{k+1}-B_k\rightarrow 0\quad \Rightarrow\quad {U^*}^TY^*V^*=0
\end{split}}
\end{equation}

\noindent and from this the conditions (\ref{KKT}) are easily derived. $\blacksquare$.
As for the convergence rate, a similar argument to the one used in \cite{ALM} shows that the convergence rate is $\mathcal{O}(1/\mu_k)$.

\section{Experimental Evaluation}
\label{sec:results}

\begin{figure}[!t]
	\centering
	\vspace{-1mm}
	\includegraphics[scale=0.64]{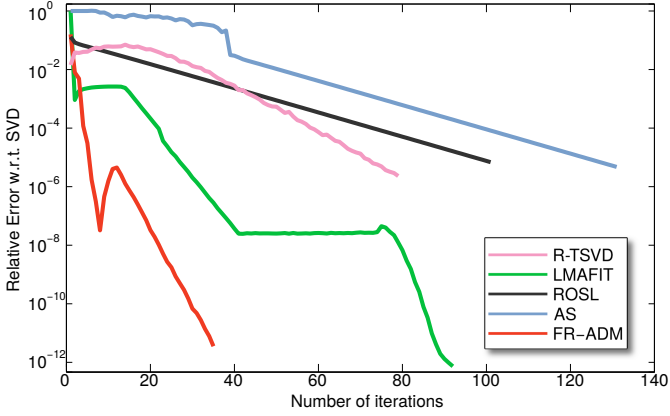}
	\vspace{-2mm}
	\caption{Convergence speed of different Fixed-rank projectors with respect to an exact TSVD.}
	\label{fig:svd}
	\vspace{5mm}
\end{figure}

\begin{figure*}[!t]
	\centering
	\vspace{-1mm}
	\includegraphics[scale=0.6]{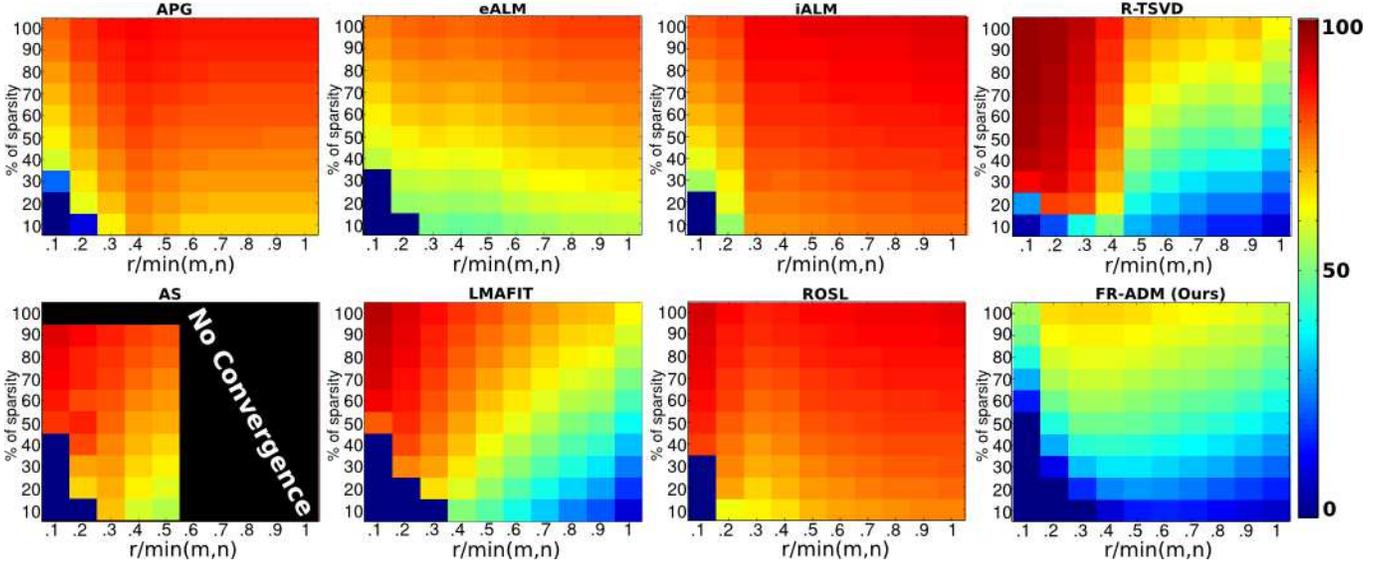}
	\vspace{-2mm}
	\caption{Phase transition diagrams for APG, eALM, iALM, , AS, LMAFIT, ROSL and FR-ADM, showing the percentage of error according to the percentage of outliers (y-axis) and the fraction of the matrix rank $r/min(m,n)$ for $m=n=800$ (x-axis). Probabilities are calculated as $\frac{1}{K} \sum_{z=1}^K \sum_{i,j}^{m,n} \frac{\psi_{\epsilon}(S^z_{i,j}-S^{*,z}_{i,j})}{m n}$, where $K$ is the number of repetitions and $\psi_{\epsilon}(s) = \{|s|, \text{ if } |s| > \epsilon;\ \ 0, \text{otherwise} \}$.}
	\label{fig:phase}
	\vspace{-3mm}
\end{figure*}

FR-ADM, and its Nystrom accelerated variant FR-Nys, are compared here against the selected methods of the state of the art, i.e.: Accelerated Proximal Gradients (APG)~\cite{APG}, inexact Augmented Lagrangian Multiplier (iALM), exact Augmented Lagrangian Multiplier (eALM)~\cite{ALM}, Active Subspace method (AS)~\cite{AS}, Low-Rank Matrix Fitting (LMAFIT)~\cite{LMAFIT} and Robust Orthonormal Subspace Learning (ROSL)~\cite{ROSL}. These methods are good representatives of the evolution of \textit{S-LR} and \textit{S-FR} solutions, ranging from the fundamental proximal gradients of APG to sophisticated factorizations included in AS and ROSL, via ADMM optimization~\cite{ALM}. We also included a version of iALM that makes use of the Randomized TSVD (R-TSVD)~\cite{Halko} in order to show the benefits of our approach against simple randomization. It is critical for the correct understanding of our experiment to clarify that we have split the previous methods up into two categories, i.e., \textit{S-LR} and \textit{S-FR} techniques. APG, iALM and eALM represent \textit{S-LR} methods, i.e., the rank is not known a priori; while R-TSVD, AS, LMAFIT, ROSL and FR-ADM represent \textit{S-FR} solutions, i.e., a correct initialization of the rank is provided, since the specific application allows it. This assumption holds for the entire section. Experiments are conducted over synthetic and real data to show the capabilities of our technique in computer vision problems. All the algorithms have been configured according to the suggestions of their respective authors. The experiments were run on a desktop PC at 3.2GHz and 64GB of RAM.

\subsection{Synthetic Data}

We test the recovery accuracy and time performance of the methods with matrices of different dimensions and ranks. To this end, we generate full-rank matrices $A \in \mathbb{R}_*^{m \times r}$ and $B \in \mathbb{R}_*^{n \times r}$ from a Gaussian distribution $N(0, 1)$, such that $L = A B^T$ and $\text{rank}(L)=r$. A sparse matrix $S \in \mathbb{R}^{m \times n}$ representing outliers is created with a given percentage of its entries being non-zero and magnitudes in the range $[-1, 1]$. Then, the final corrupted matrix is $M = L + S$. We deliberately forced the sparse entries to have a magnitude similar to the one of the expected low-rank matrix. The reason for this is that usually the experiments presented in the literature impose a good differentiation between the magnitude of the entries of $L$ and $S$, making the recovery problem almost trivial. Here, we remove that simplification, allowing for similar magnitudes of the corrupted entries, which makes the problem more interesting. We will show that, with this challenging setup, the performance of many state-of-the-art methods dramatically decreases, while our approach maintains a good recovery accuracy. 

Our first test measures the recovery capabilities of the different methods under study when subjected to similar magnitudes of the entries of $L$ and $S$. To this end we create corrupted matrices of increasing rank and an increasing fraction of outliers. The result of this experiment is shown in Figure~\ref{fig:phase} in form of phase transition diagrams, with rank fractions represented in the x-axis and outlier fraction in the y-axis. Colors represent the recovery (inverse) probability of each case, i.e., the lower error (cold colors, i.e. blue-ish) the better. From this plot, it can be seen that these conditions are very challenging for all the algorithms.

APG, eALM and iALM, making use of an SVD, end up with a very narrow recovery region (in blue). R-TSVD gets a narrow recovery region due to accuracy problems (see also Fig.~\ref{fig:svd} for further information). Notice that AS is not even able to converge beyond a $60\%$ of rank due to the strong non-convexity induced by its bi-linear factorization. LMAFIT shows a rather acceptable recovery region, while ROSL clearly suffers in obtaining a correct recovery for this sort of data. In our analysis, ROSL performs well when the magnitude of $S$ (the noisy entries) are several magnitudes bigger than those of $L$. However, in other cases the recoverability of ROSL dramatically decay. Our proposal, FR-ADM, presents the best recovery for a wider region even in this challenging setup. This characteristic is critical for real applications where outliers might be either very large or very subtle. 

We also evaluated one of the most critical aspects of these methods, i.e., the accuracy of a given method at providing a good low-rank approximations of a matrix $L$. State-of-the-art approaches have gained in efficiency by replacing the SVD for a more convenient fixed-rank projection, as in the case of R-TSVD, AS, LMAFIT, ROSL and our proposal FR-ADM. However, as shown in Figure~\ref{fig:svd}, different projection strategies lead to different convergence rates and speeds. In this way, when compared against an exact TSVD, the polar decomposition used by FR-ADM turns out to be superior to all its competitors, as derived from the reduced number of iterations required to achieve a relative error of $10^{-12}$. We would like to highlight that our approach even presents a better convergence behaviour than the well-known R-TSVD, which is considered one of the fastest methods for low-rank projection. Later, we will show that FR-ADM not only has a better convergence, but is also faster and more accurate.

Our second experiment uses matrices of increasing sizes ($m=n$), ranging from $m=500$ to $m=8000$, while keeping the rank fixed, $r = 10$ and the entries magnitudes as defined above. 10 repetitions are considered per each size. In this case the methods under evaluation are the APG, iALM, eALM, R-TSVD, AS, LMAFIT, ROSL, ROSL+ and our proposal FR-ADM, along with its equivalent accelerated version, FR-Nys. We have accelerated FR-ADM to present a counterpart to ROSL+\cite{ROSL}. In this way we can offer a fair comparison with our proposal and show that our method remains superior after the Nystrom's speed-up. Results of this test are shown in Table~\ref{tab:rank10}, considering the recovery error for both matrices $L$ and $S$ given by $\text{Err. L} = \norm{L-L*}_F / \norm{L*}_F$ and $\text{Err. S} = \norm{S-S*}_F / \norm{S*}_F$, where $L*$ and $S*$ are the optimal matrices. We also consider computational time in seconds and the number of iterations used by each method. FR-ADM is the method with the best trade-off of high recovery accuracy and low computational time (the fastest of the non-accelerated methods). The efficiency of methods such as LMAFIT and ROSL considerably decreased due to their difficulties to face small sparse entries. APG, iALM and eALM also find troubles searching for the appropriate rank in this challenging conditions. For the case of the R-TSVD, its accuracy is lower than desired, and due to its lack of accuracy requires too many iterations to converge.

For the accelerated methods, FR-Nys has proven to be the fastest and the most accurate in all synthetic tests, despite the accuracy degradation provoked by the matrix sampling. The benefits of applying Nystrom's acceleration are clear, specially for big matrices, as in the $8000 \times 8000$ case, where the total time is reduced in two orders of magnitude. However, as we show in the next experiment, this acceleration is not convenient for problems with large matrix ranks. 

In this third experiment methods performance is tested against matrices of increasing dimensions and rank. Matrices are created as described above, but their rank is established to be $rank(L) = 0.1 m$, where $m=n$ is the matrix size. Results are summarized in Table~\ref{tab:bigRank}. The first thing to notice is that the time of Nystrom-accelerated methods is bigger than their unaccelerated counterparts.  This is due to the high rank of the problem and that the matrices resulting from the Nystrom's sampling technique are of sizes $m \times k r$ and $k r \times n$, with $k$ big enough (usually $3 < k < 10$). This leads to two matrices that are almost of the size of the original one, making the use of Nystrom counterproductive. Regarding the unaccelerated methods, FR-ADM performs almost twice faster than the second best approach, AS. In terms of recovery accuracy, all the methods present similar results, except for small matrices, where ROSL fails due to its sensibility to initialization parameters.

\singlespacing
\begin{table}
\centering
\resizebox{\columnwidth}{!}{%
\scriptsize
\tabcolsep=0.06cm
\begin{tabular}{|c|c|c|c|c|c|c|c|c|c|c|c|}
\cline{3-12}\noalign{\vskip 1pt}
\multicolumn{2}{c|}{ } & \multicolumn{3}{|c|}{\textit{S-LR}} & \multicolumn{5}{|c|}{\textit{S-FR} No Accel.} & \multicolumn{2}{|c|}{\textit{S-FR} Accel. }\\
\cline{3-12}\noalign{\vskip 1pt}
\multicolumn{2}{c|}{ } & \textbf{APG} & \textbf{iALM} & \textbf{eALM} & \textbf{R-TSVD} & \textbf{AS} & \textbf{LMAFIT} & \textbf{ROSL} & \textbf{FR-ADM} & \textbf{ROSL+} & \textbf{FR-Nys}\\
 \hline
				& Err. L & 6.2e-7	& 6.3e-10 	&  3.3e-9	& 1.0e-7		& 7.6e-9		& 1.3e-4			& 5.0e-10	& 2.21e-10		& 1.7e-2		& 6.8e-10\\
				& Err. S & 8.4e-5	& 1.1e-7		&  5.8e-7	& 8.7e-6		& 3.8e-7		& 9.1e-3			& 7.5e-8		& 3.6e-8			& 1.2e+0		& 4.8e-8\\
500				& iters  & 140		& 33			&  9			& 96			& 120		& 40				& 93			& 28				& 200		& 68\\
				& time   & 11.74		& 1.25		&  2.29		& 0.90		& 0.59		&\colB{0.11}		& 0.89		&\colB{0.11}		& 0.67		& \colR{0.17}\\
				\hline
				& Err. L & 4.7e-7	& 1.1e-9		& 2.5e-10	& 1.8e-7		& 1.4e-9		& 1.8e-7			& 1.3e-9		& 5.0e-10		& 1.2e-4		& 6.2e-10\\
				& Err. S & 8.7e-5	& 5.1e-7		& 6.4e-8		& 1.6e-5		& 4.3e-7		& 1.2e-5			& 6.3e-7		& 2.0e-7			& 8.4e-3		& 8.7e-7\\
1K				& iters	 & 142		& 34			& 10			& 95			& 133		& 65				& 98			& 28				& 200		& 65\\
				& time	 & 61.75		& 5.87		& 11.04		& 1.57		& 2.39		& 0.75			& 4.27		& \colB{0.46}	& 1.29		& \colR{0.22}\\
				\hline
				& Err. L & 3.7e-7	& 8.1e-10	& 2.3e-10	& 4.4e-7		& 1.3e-9		& 2.1e-8			& 1.1e-9		& 3.0e-10		& 5.2e-8		& 3.1e-10\\
				& Err. S & 9.4e-5	& 4.5e-7		& 7.8e-8		& 3.4e-5		& 6.3e-7		& 4.6e-7			& 6.9e-7		& 1.8e-7			& 3.6e-6		& 7.6e-8\\
2K				& iters  & 144		& 35			& 10			& 92			& 131		& 300			& 98			& 29				& 200		& 66\\
				& time   & 396.4		& 20.34		& 50.02		& 5.14		& 9.64		& 12.45			& 17.79		& \colB{1.57}	& 1.98		& \colR{0.31}\\
				\hline
				& Err. L & 2.6e-7	& 4.8e-10	& 2.5e-10	& 7.3e-7		& 9.4e-10	& 1.3e-8			& 6.8e-10	& 2.7e-10		& 4.7e-9		& 3.3e-10\\
				& Err. S & 9.3e-5	& 4.2e-7		& 1.0e-7		& 5.9e-5		& 6.3e-7		& 2.9e-7			& 6.0e-7		& 2.2e-7			& 3.2e-7		& 2.2e-8\\
4K				& iters  & 147		& 36			& 10			& 91			& 135		& 300			& 99			& 29				& 140		& 62\\
				& time   & 3002		& 112		& 328		& 18			& 50.05		& 56.93			& 67.63		& \colB{7.52}	& 3.33		& \colR{0.65}\\
				\hline 
				& Err. L & 1.9e-7	& 5.4e-10	& 2.7e-10	& 1.4e-6		& 6.7e-10	& 9.4e-9			& 4.9e-10	& 4.5e-10		& 5.4e-9		& 5.0e-10\\
				& Err. S & 9.3e-5	& 6.5e-7		& 1.6e-7		& 1.3e-4		& 6.8e-7		& 2.0e-7			& 6.2e-7		& 7.6e-7			& 3.7e-7		& 2.8e-8\\
8K				& iters  & 150		& 36			& 10			& 88			& 138		& 300			& 100		& 28				& 139		& 61\\
				& time   & 22415		& 517		& 2214		& 63.6		& 243		& 202			& 261		& \colB{27.42}	& 6.90		& \colR{1.24}\\
				\hline

\end{tabular}%
}
\vspace{0.7mm}
\caption{Average evaluation of recovery accuracy and computational performance for matrices of different dimensions, with  10\% outliers and $\text{rank}(L) = 10$ across ten repetitions. Best time of an accelerated method is shown in \colR{red} and the best time of an unaccelerated method is shown in \colB{blue}.}
\label{tab:rank10} 
\vspace{1mm}
\end{table}

\singlespacing
\begin{table}
\centering
\resizebox{\columnwidth}{!}{%
\scriptsize
\tabcolsep=0.06cm
\begin{tabular}{|c|c|c|c|c|c|c|c|c|c|c|c|}
\cline{3-12}\noalign{\vskip 1pt}
\multicolumn{2}{c|}{ } & \multicolumn{3}{|c|}{\textit{S-LR}} & \multicolumn{5}{|c|}{\textit{S-FR} No Accel.} & \multicolumn{2}{|c|}{\textit{S-FR} Accel. }\\
\cline{3-12}\noalign{\vskip 1pt}\multicolumn{2}{c|}{ } & \textbf{APG} & \textbf{iALM} & \textbf{eALM} & \textbf{R-TSVD} & \textbf{AS} & \textbf{LMAFIT} & \textbf{ROSL} & \textbf{FR-ADM} & \textbf{ROSL+} & \textbf{FR-Nys}\\
\hline
				& Err. L & 1.3e-5	& 1.7e-4		& 1.1e-6		& 2.3e-7		& 1.5e-8		& 1.8e-4		& 1.2e-2		& 9.0e-9			& 1.7e-4		& 2.7e-10\\
500				& Err. S & 1.3e-4	& 1.7e-3		& 1.1e-5		& 4.1e-5		& 1.1e-7		& 2.8e-2		& 1.2e-1		& 1.1e-7			& 2.5e-2		& 4.1e-8\\
r=50				& iters  & 175		& 37			& 11			& 88			& 85			& 48			& 138		& 37				& 200		& 66\\
				& time   & 13.77		& 3.61		& 15.44		& 1.70		& 0.75		& 0.42		& 6.43		&\colB{0.34}		& 8.99		& \colR{0.97}\\
				\hline   
				& Err. L & 2.4e-6	& 6.9e-9		& 4.4e-7		& 6.7e-7		& 1.6e-8		& 2.8e-8		& 1.2e-8		& 9.9e-9			& 4.3e-8		& 8.1e-10\\
1K				& Err. S & 3.4e-5	& 1.3e-7		& 6.2e-6		& 1.6e-4		& 1.7e-7		& 2.2e-6		& 8.8e-8		& 1.8e-7			& 9.4e-6		& 1.8e-7\\
r=100			& iters  & 174		& 37			& 11			& 82			& 79			& 300		& 69			& 36				& 200		& 62\\
				& time   & 68.88		& 14.36		& 61.74		& 7.63		& 3.18		& 7.43		& 33.13		&\colB{1.54}		& 69.75		& \colR{3.61}\\
				\hline
				& Err. L & 6.8e-7	& 6.3e-9		& 2.0e-7		& 1.7e-6		& 1.1e-8		& 1.9e-8		& 1.1e-8		& 1.1e-8			& 4.5e-8		& 1.6e-9\\
2K				& Err. S & 1.3e-5	& 1.7e-7		& 4.0e-6		& 5.9e-4		& 1.6e-7		& 2.0e-6		& 1.2e-7		& 3.0e-7			& 1.4e-5		& 4.8e-7\\	
r=200			& iters  & 175		& 37			& 11			& 77			& 85			& 300		& 67			& 35				& 200		& 60\\
				& time   & 461		& 80.49		& 332		& 32.74		& 17.32		& 30.77		& 325		&\colB{7.53}		& 653.13		& \colR{16.93}\\
				\hline
				& Err. L & 3.5e-7	& 7.5e-9		& 1.5e-7		& 5.0e-6		& 1.2e-8		& 1.3e-8		& 1.0e-8		& 6.8e-9			& 4.7e-8		& 5.3e-9\\
4K				& Err. S & 1.0e-5	& 2.8e-7		& 4.4e-6		& 2.4e-3		& 2.7e-7		& 1.9e-6		& 1.8e-7		& 2.6e-7			& 2.0e-5		& 2.3e-6\\
r=400			& iters  & 175		& 37			& 10			& 71			& 89			& 300		& 66			& 36				& 200		& 57\\
				& time   & 3453		& 529		& 2106		& 183.5		& 107		& 162		& 3586		&\colB{43}		& 7008		& \colR{96.57}\\
				\hline
				& Err. L & 5.9e-7	& 5.0e-9		& 4.3e-9		& 1.5e-5		& 6.3e-9		& 8.6e-9		& 7.1e-8		& 9.5e-9			& 4.8e-8		& 1.4e-8\\
8K				& Err. S & 3.7e-4	& 5.5e-6		& 3.4e-6		& 1.0e-2		& 6.6e-6		& 1.8e-6		& 1.2e-5		& 1e-5			& 3.0e-5		& 8.6e-6\\
r=800			& iters  & 143		& 35			& 10			& 65			& 130		& 300		& 7107		& 21				& 200		& 54\\
				& time   & 23130		& 2651		& 6394		& 1382		& 1075		& 1035		& 97397		& \colB{166}		& 91242		& \colR{564}\\
				\hline
\end{tabular}%
}
\vspace{0.7mm}
\caption{Average evaluation of recovery accuracy and computational performance for matrices of different dimensions, with  10\% outliers and $\text{rank}(L) = 0.1 m$ across ten repetitions. Best time of an accelerated method is shown in \colR{red} and the best time of an unaccelerated method is shown in \colB{blue}.}
\label{tab:bigRank} 
\vspace{3mm}
\end{table}

\vspace{-8mm}
\subsection{Robust Photometric Stereo}
\vspace{-1mm}
We have chosen photometric stereo~\cite{PhotoStereo} (PS) as our first example of fixed-rank problem. PS consists in estimating the normal map and depth of a still scene from several 2D images grabbed from the same position but under different light directions. The Lambertian reflectance model~\cite{PhotoStereo} is assumed, such that the light directions $L \in \mathbb{R}^{3 \times n}$, the matrix of normals (unknowns) $N \in \mathbb{R}^{m \times 3}$, and the matrix of pixel intensities $I \in \mathbb{R}^{m \times n}$ are related via $I = \rho N L$, where $\rho$ represents the albedo. The objective of recovering the normal map $N$ can be achieved by a Least-Squares (LS) method, but the quality of such a solution would suffer in the presence of outliers. Instead, robust decompositions can be used to get ride of outliers, as proposed in~\cite{RobustPS}. Since $I$ is a product of two rank-3 matrices, in ideal conditions its rank is at most 3. We make use of this rank property to recover an uncorrupted version of $I$ that leads to a better estimation of the map $N$ and consequently of the depth map.

In our tests we use a dataset of objects viewed under 20 different illuminations, provided in ~\cite{PS}. From such images, we recover an uncorrupted version of the intensities $I$. Then we run the Photometric Stereo Toolbox~\cite{PS} to recover normal maps, depth maps, 3D models and some statistics. Table~\ref{tab:ps1} shows the error in the normal maps after the recovery process with different methods. Here, we consider the reconstruction error, i.e., the normal map is re-rendered into a shading image and then compared with the captured images. From the resulting error map several statistics are computed (RMS, mean and maximum error). The classical LS approach is taken as a reference of non-robust approaches. As robust methods, APG, eALM and iALM, AS, LMAFIT, ROSL and FR-ADM are considered. R-TSVD has not been considered due to its observed reduced accuracy.  Nystrom accelerated versions are excluded due to the small size of the observation matrices, a constraint that prevents speed-ups.

\begin{figure*}[!t]
	\centering
	\vspace{-1mm}
	\includegraphics[scale=0.76]{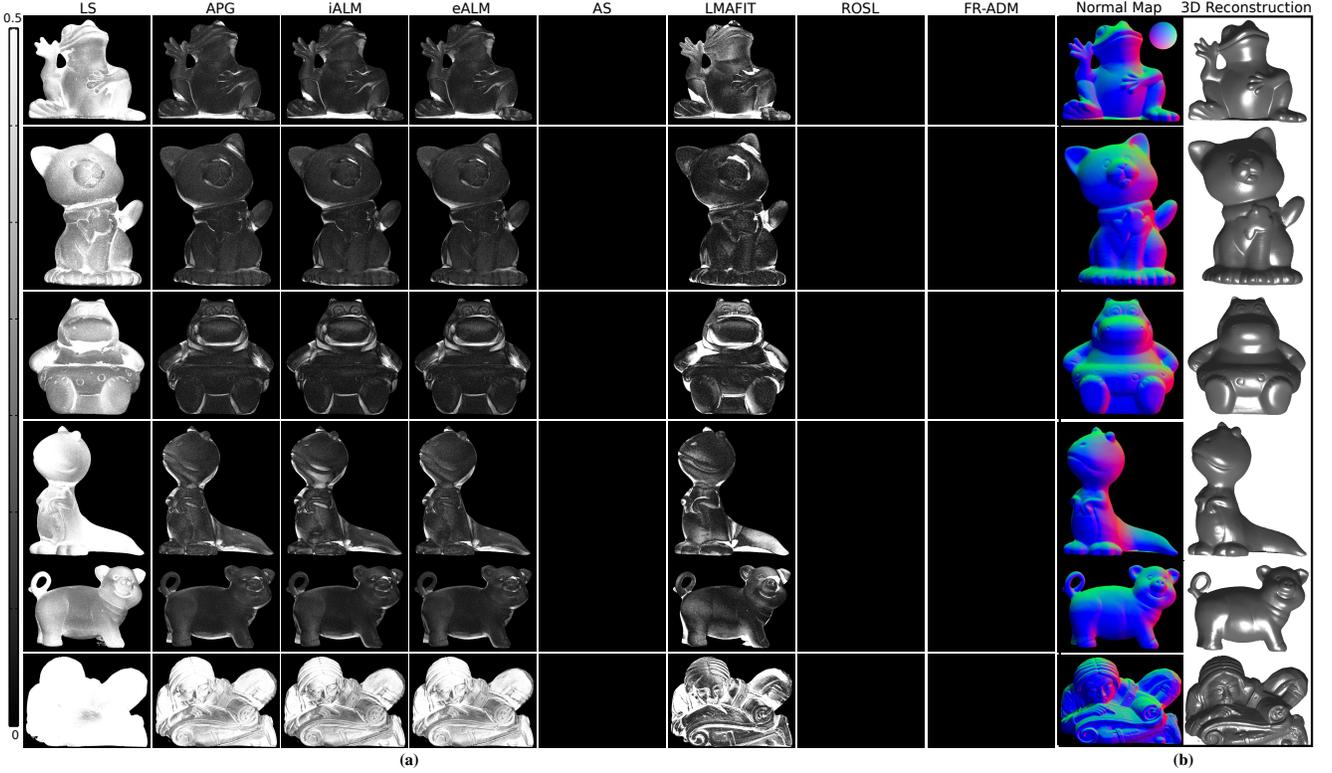}
	\vspace{-2mm}
	\caption{\textbf{(a)} Normal error maps after the reconstruction, with intensities scaled by $100$ for visualization. Notice that the errors of AS, ROSL and FR-ADM are insignificant, below $10^{-10}$. \textbf{(b)} 3D reconstruction of the objects after the application of the FR-ADM technique.}
	\label{fig:psComb}
	\vspace{-3mm}
\end{figure*}

The comparison shows that AS, ROSL and FR-ADM are the most accurate methods, producing estimations of the normal map with reconstruction errors below $10^{-10}$. The remaining methods are far from the accuracy offered by these fixed-rank techniques, producing high residuals. Although AS consistently presents a lower error in the majority of the cases, the error differences below $10^{-10}$ are of no impact for the application. This is shown in the error maps of Fig.~\ref{fig:psComb}\textbf{(a)}. However, computational time is a critical factor for this problem, where, FR-ADM is one order of magnitude faster than ROSL and two orders faster than AS. 

This figure displays the error maps of the considered approaches. As expected, LS leads to high errors due to outliers. APG, iALM and eALM improve LS results, but since they do not use the rank-$3$ constraint recovered matrices have an erroneous low-rank. Fixed-rank techniques, such as AS, ROSL and FR-ADM achieve very low residuals, making the error maps black. The recovered normal maps after the application of the FR-ADM technique are shown in Fig.~\ref{fig:psComb}\textbf{(b)} along with the 3D reconstruction of the objects. It can be concluded that \textit{S-FR} techniques, can drastically benefit problems like photometric stereo and FR-ADM stands as the fastest alternative while offering a very high accuracy.

\singlespacing
\begin{table}
\resizebox{\columnwidth}{!}{%
\scriptsize
\tabcolsep=0.07cm
\begin{tabular}{|c|c|c|c|c|c|c|c|c|c|}
\cline{3-10}\noalign{\vskip 1pt}
\multicolumn{2}{c|}{ } & LS & APG & iALM & eALM & AS  & ROSL & LMAFIT & FR-ADM\\
 \hline
		&\tx{RMS} 		&	1.4e-2 	&	3.7e-3 		& 3.9e-3 	& 3.9e-3 	& 1.2e-12	& 1.6e-11 	& 2.3e-2 	& 1.5e-11  \\
\ty Frog		&\tx{Mean Err.}	& 	1.1e-2 	&	2.7e-3 		& 2.7e-3 	& 2.7e-3 	& 1.2e-12  	& 1.4e-11 	& 7.9e-3 	& 1.3e-11  \\
		&\tx{Max Err.}	&	1.6e-1 	&	2.2e-2 		& 2.1e-2 	& 2.1e-2 	& 1.8e-12  	& 4.8e-11 	& 2.1e-1 	& 4.7e-11  \\
		& \tx{Time(s)}	& 	x 		&	2.3e+2 		& 1.4e+2 	& 5.6e+2		& 3.1e+1 	& 4.0e+1 	& 1.4e+2 	& \colB{7.1e+0} \\
\hline
 		&\tx{RMS} 		&	1.4e-2 	&	2.7e-3 		& 2.5e-3 	& 2.5e-3		& 4.3e-14 	& 2.7e-11 	& 9.6e-3 	& 2.5e-11  \\ 
\ty Cat		&\tx{Mean Err.}	&	9.3e-3 	&	1.9e-3 		& 1.8e-3 	& 1.8e-3		& 4.1e-14 	& 2.3e-11 	& 3.9e-3 	& 2.2e-11  \\
		&\tx{Max Err.} 	&	2.2e-1 	&	1.8e-2 		& 1.4e-2 	& 1.4e-2		& 6.4e-14 	& 6.6e-11 	& 1.4e-1 	& 6.7e-11 \\
		&\tx{Time(s)} 	&	x 		&	1.8e+2 		& 1.1e+2 	& 4.3e+2		& 2.4e+1 	& 3.0e+1 	& 1.1e+2 	& \colB{5.9e+0}  \\
\hline
		&\tx{RMS} 		&	1.5e-2 	&	2.9e-3 		& 2.8e-3 	& 2.8e-3		& 6.0e-13 	& 2.6e-11 	& 1.4e-2 	& 2.6e-11 \\
\ty Hippo	&\tx{Mean Err.}	&	9.8e-3 	&	1.6e-3 		& 1.5e-3 	& 1.5e-3		& 5.7e-13  	& 2.4e-11 	& 6.4e-3 	& 2.3e-11 \\
		&\tx{Max Err.} 	&	1.9e-1 	&	2.3e-2 		& 1.9e-2 	& 1.9e-2		& 9.8e-13 	& 8.1e-11 	& 1.8e-1 	& 8.4e-11 \\
		&\tx{Time(s)} 	& 	x		&	1.9e+2 		& 1.2e+2		& 4.7e+2		& 2.6e+1 	& 3.2e+1 	& 1.2e+2 	& \colB{6.0e+0} \\
\hline
		&\tx{RMS} 		&	1.4e-2 	&	4.0e-3 		& 3.9e-3 	& 3.6e-3		& 3.8e-12 	& 1.8e-11 	& 1.8e-2 	& 1.5e-11 \\
\ty Lizard	&\tx{Mean Err.}	&	1.2e-2 	&	3.1e-3 		& 3.0e-3 	& 2.8e-3		& 3.5e-12 	& 1.6e-11 	& 6.2e-3  	& 1.3e-11 \\
		&\tx{Max Err.}	&	1.7e-1 	&	3.6e-2 		& 2.7e-2 	& 2.7e-2		& 1.2e-11 	& 5.5e-11 	& 2.2e-1 	& 4.4e-11 \\
		&\tx{Time(s)	}	& 	x 		&	2.8e+2 		& 1.6e+2		& 7.8e+2		& 3.7e+1 	& 4.3e+1 	& 1.6e+2 	& \colB{8.9e+0} \\
\hline
		&\tx{RMS}		&	1.0e-2 	&	2.7e-3 		& 2.5e-3 	& 2.5e-3		& 1.4e-11 	& 1.9e-14 	& 1.5e-2 	& 6.8e-11 \\
\ty Pig		&\tx{Mean Err.}	&	7.9e-3 	&	2.2e-3 		& 2.1e-3 	& 2.0e-3		& 1.4e-11 	& 1.5e-14 	& 5.1e-3 	& 5.5e-11 \\
		&\tx{Max Err}	&	2.1e-1 	&	1.2e-2 		& 1.5e-2 	& 1.4e-2		& 2.7e-11 	& 8.7e-14	& 2.2e-1  	& 3.1e-10 \\
		&\tx{Time(s)}	& 	x 		&	2.3e+2		& 1.4e+2		& 5.2e+2		& 3.2e+1 	& 3.7e+1		& 1.5e+2 	& \colB{7.7e+0} \\
\hline
		&\tx{RMS} 		&	4.3e-2 	&	1.1e-2 		& 9.1e-3 	& 9.9e-3		& 8.8e-13 	& 2.8e-13	& 2.7e-2 	& {1.3e-13}\\
\ty Scholar &\tx{Mean Err.}	& 	3.3e-2 	&	1.0e-2 		& 8.4e-3 	& 9.2e-3		& 7.9e-13 	& 2.2e-13	& 1.5e-2 	& {1.0e-13} \\
		&\tx{Max Err.}	&	3.3e-1 	&	3.3e-2		& 2.2e-2 	& 2.4e-2		& 1.9e-12 	& 1.3e-12	& 2.4e-1  	& {6.0e-13} \\
		&\tx{Time(s)}	& 	x 		&	5.0e+2		& 3.0e+2		& 1.3e+3		& 6.5e+1 	& 8.0e+1		& 3.1e+2 	& \colB{1.5e+1} \\
\hline
\end{tabular}%
}
\vspace{0.7mm}
\caption{Evaluation of the reconstruction error for the photometric stereo dataset~\cite{PS}. The time taken for the LS method is not included in the evaluation.}
\label{tab:ps1} 
\vspace{3mm}
\end{table}

\vspace{-3mm}
\subsection{Robust Spectral Clustering}
\begin{figure}[!t]
	\centering
	\vspace{0mm}
	\includegraphics[scale=0.62]{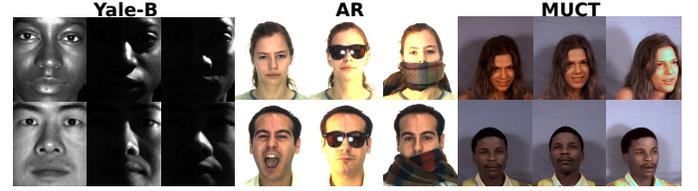}
	\caption{Instances of males and females subjects of the different data sets used in our evaluation.}
	\label{fig:clusters}
	\vspace{6mm}
\end{figure}

\vspace{-1mm}
We address clustering as a fixed-rank optimization problem with a known number of clusters represented by the matrix rank, where such a rank can become very high. Here, \textit{S-FR} methods can be easily added to the pipeline of Spectral Clustering approaches (SP)\cite{PSCDS} to increase robustness to outliers and improve accuracy.  We consider the problem of clustering faces given the number of categories for three face data sets, i.e., the Extended Yale Face Database B~\cite{YaleB} (16128 images of 38 different subjects), the AR Face database~\cite{AR} (4000 images of 126 different subjects) and the MUCT Face Database~\cite{MUCT} (3755 images of 625 different subjects). All of them contain people under different illumination conditions. In addition, MUCT and AR include pose variations, and in the case of AR people use different outfits (see Fig.\ref{fig:clusters} for some examples).


In our experiments we use the Parallel Spectral Clustering in Distributed Systems (PSCDS)~\cite{PSCDS} method as the base code for spectral clustering, but just employ a simple desktop machine. The different \textit{S-FR} methods are incorporated to PSCDS as a preprocessing stage as follows. First, each image is described by the Gist~\cite{Gist} holistic descriptor with $5$ scales of $8$ orientations and $12$ blocks. This produces a vector of $5760$ dimensions. The use of Gist instead of the original images has consistently produced an improvement in accuracy in the range of [15\%, 20\%]. Secondly, all the descriptors of a dataset are combined forming an observation matrix $A = N \times 5760$, where $N$ is the total number of images in the specific dataset. The rank of $A$ is the number of expected clusters $C_{\text{rank}}$. Then, the \textit{S-FR} method under evaluation recovers a subspace $U_A$ of rank $C_{\text{rank}}$ from $A$. The matrix $U_A$ is then used in the pipeline of  PSCDS to compute the distance matrix $W_U$, considering five nearest neighbours per sample, followed by the spectral clustering.

We have considered LMAFIT, ROSL, ROSL+, FR-ADM and FR-Nys as representatives of the \textit{S-FR} approaches. Additionally, we also compared against state-of-the-art clustering techniques such as the Robust Subspace Segmentation by Low-Rank Representation (LRR)~\cite{LRR} and the Smooth Representation Clustering (SMR)~\cite{SMR}, specifically designed for clustering purposes. The results of our evaluation are presented in Table~\ref{tab:clustering}, including the average clustering error (ce); the base time, i.e., time taken by the specific \textit{S-FR} method; and the total time, i.e., base time plus the time taken by the PSCDS. For LRR and SMR the total time is that produced by the method.

When considering the Yale-B and AR datasets FR-ADM obtains the lowest clustering errors, 2.7\% and 6.65\% respectively. Moreover, FR-ADM and FR-Nys present the best balance between accuracy and computational time for these datasets.  MUCT, is the most challenging dataset with 625 classes, which is a very high rank in comparison to its matrix dimensions ($3755 \times 5760$). These conditions are beyond the recovery boundaries of \textit{S-FR} methods, and even though FR-ADM accuracy is comparable to that obtained by the top method, LRR. Furthermore, FR-ADM computational performance is more than 20 times faster than LRR for this case, supporting the good accuracy-speed trade-off offered by the method. 


\singlespacing
\begin{table}
\resizebox{\columnwidth}{!}{%
\scriptsize
\tabcolsep=0.11cm
\begin{tabular}{|c|c|c|c|c|c|c|c|c|c|}
\cline{3-10}\noalign{\vskip 1pt}
\multicolumn{2}{c|}{ } & PSCDS & LRR & SMR & LMAFIT & ROSL & ROSL+ & FR-ADM & FR-Nys\\
\hline
Yale-B	   & ce (\%)			& 18.7\%		& 13.8\%		& 28.4\% 	& 18.8\%		& 20.1\%  & 30.4\% 	& 2.7\%		& 2.89\%\\
A=16128x5760 & base time		& 5.17		& 64.8		& 351.6		& 2.4		& 274.6	  &	7.6    	& 6.8		& 0.58\\
$C_{\text{rank}}$=38	   & total time		& 5.17		& 64.8		& 351.6		& 5.6		& 275.5	  & 8.7		& 8.8		& \colB{2.5} \\
	   \hline
AR	   & ce (\%)			& 17.2\%		& 36.8\%		& 39.7\%		& 6.70\%		& 7.17\%	  & 46.0\%	& 6.65\%		& 7.17\%		\\
A=4000x5760   & base time 	   	& 5.08  		& 606.8		& 105.1		& 17.6		& 662.2	  & 48.1		& 13.81		& 1.63		\\
$C_{\text{rank}}$=126	   & total time		& 5.08		& 606.8		& 105.1		& 21.05		& 665.1	  & 49.9		& 16.91		& \colB{3.83}		\\
	   \hline
MUCT	   & ce (\%)			& 55.3\%		& 53.4\%		& 55.8\%		& 56.2\%		& 56.3\%	  &	78.4\%	& 55.7\%		& 62.8\%		\\
A=3755x5760	   & base time		& 76.7		& 3820		& 3995		& 101.2		& 17696	  &	4890		& 85.5		& 67.2		\\
$C_{\text{rank}}$=625	   & total time		& \colB{76.7}		& 3820		& 3995		& 190.9		& 17771	  &	4977		& 175.2		& 156.3		\\
	   \hline
\end{tabular}%
}
\vspace{0.7mm}
\caption{Clustering errors including time evaluation. Base time refers to the time used by the specific \textit{S-FR} method, while total time refers to the time required to perform the full clustering task.}
\label{tab:clustering} 
\vspace{3mm}
\end{table}

\vspace{-3mm}
\section{Conclusion and Future Work}
\label{sec:conclusion}
\vspace{-1mm}

In this paper we have proposed an efficient, stable and accurate technique, FR-ADM, to perform a robust decomposition of a corrupted matrix into its fixed-rank and sparse components. To this end we have based our algorithm on a polar factorization on a product manifold $(St \times SPD \times St)/O_r$, combining key tools from manifold optimization and fast projectors. We also proposed a fast $\text{SPD}$ projector to speed up computation, along with a proof of its validity in this context.  Additionally, Nystrom's sampling techniques have been used to further accelerate the results, achieving a linear complexity. The resulting algorithm has been tested on synthetic cases and the challenging problems of robust photometric stereo and spectral clustering, proving to be as accurate and more efficient than state-of-the-art approaches and paving the way towards large-scale problems.
\section*{Appendix: Convergence analysis of the FixedRankOptFull algorithm}
\label{appendix}

In this Appendix we shall show that the minimization subproblem (\ref{eq:MinimizationFixedRank}), i.e.
\begin{equation}
\textrm{min}_{U\in \textrm{St}_{m,r},B\in \textrm{SPD}_r,V\in \textrm{St}_{n,r}}\norm{M-UB V^T}_F^2\,,
\end{equation}
although highly non-convex, converges geometrically to the global minimum when optimized via the proposed 
\textit{FixedRankOptFull} method (Algorithm~\ref{Alg-FROpt}). 

\begin{algorithm}
 \algsetup{linenosize=\small}
  \small
\begin{algorithmic}[1]
        \REQUIRE Data matrix $M\in \mathbb{R}^{m\times n}$, initial matrices $U_0\in \textrm{St}_{m,r}$, $B_0\in \textrm{SPD}_r$, $V_0\in \textrm{St}_{n,r}$
	\STATE $i\leftarrow 0$
	\WHILE{not converged}
	\STATE $(U_{i+1},B_{i+1},V_{i+1})\leftarrow FixedRankOptStep(M,U_i,B_i,V_i)$
	\STATE $i\leftarrow i+1$
	\ENDWHILE
	\RETURN{$U^*\in \textrm{St}_{m,r}$, $B^*\in \textrm{SPD}_r$, $V^*\in \textrm{St}_{n,r}$ such that $L=U^*B^*{V^*}^T$ is the 
	TSVD of $M$ }
\end{algorithmic}
\caption{\textit{FixedRankOptFull} Algorithm}
\label{Alg-FROpt}
\end{algorithm}
\setlength{\floatsep}{0pt}
\setlength{\textfloatsep}{2mm}
\setlength{\intextsep}{2pt}

The \textit{FixedRankOptFull} algorithm performs  an alternating directions minimization (ADM)
on each of the submanifolds $\textrm{St}_{m,r}$, $\textrm{St}_{n,r}$ and $\textrm{SPD}_r$ (Algorithm~\ref{Alg-FRStep}).
In each iteration it uses the algorithm \textit{FixedRankOptStep}, described in Sec. \ref{sec:polar}, that performs a single step of the alternating directions minimization.

In Sec. \ref{sec:polar} we provided the exact projectors on each of the submanifolds 
$\textrm{St}_{m,r}$, $\textrm{St}_{n,r}$ and $\textrm{SPD}_r$, and proved the validity of the ones corresponding to the Stiefel manifolds. For the case of the $\textrm{SPD}_r$ manifold, a careful analysis is required to prove its validity.

Given that $\text{rank}(M) \geq r = \text{rank}(L)$, and considering $U$ and $V$ as the solutions of an OPP, then a unique solution in the SPD manifolds must exist. This solution is given in the following discussion, but we need some previous results.
\begin{lem}
(see \cite{Stiefel-Procrustes}) Let $\bar{U}$ and $\bar{V}$ be the solutions given by Algorithm \ref{Alg-FRStep}. Suppose $\text{rank}(B_0)=\text{rank}(MV_0)=r$, then $\bar{U}^TMV_0B_0$ and $B_0\bar{U}^TM\bar{V}$ are in $\textrm{SPD}_r$.
\label{lemmaOPP}
\end{lem}
%
\noindent \textit{Proof}: 
Since $\bar{U}\!=\!P_{\textrm{O}}[MV_0B_0]$, if $MV_0B_0\!=\!Q\Sigma S^T$, then $\bar{U}\!=\!QS^T$
and therefore $\bar{U}^TMV_0B_0$ $=SQ^TQ\Sigma S^T$ $=S\Sigma S^T$, which is SPD$_r$ since
$\text{rank}(MV_0)= \text{rank}(B_0)=r$. A similar argument, but without any additional assumption on $\text{rank}(M^T\bar{U})$, since $\text{rank}(M^T\bar{U})=\text{rank}(M^TMV_0)=\text{rank}(MV_0)=r$, shows that $\bar{V}^TM^T\bar{U}B_0$ $=B_0\bar{U}^TM\bar{V} \in \textrm{SPD}_r$ after
minimizing with respect to $V$.$\blacksquare$


Note that in Lemma~\ref{lemmaOPP} it is not neccesary that $B_0\in\text{SPD}_r$, only that it is invertible. 
Also, we conclude that $\text{rank}(\bar{U}^TM\bar{V})=r$. 
$\bar{U}^TM\bar{V}$, is in general not symmetric, although it can be written as a product of two SPD matrices, 
and therefore has positive eigenvalues.
Even though from Lemma~\ref{lemmaOPP} we have that 
$\bar{U}^TMV_0B_0$ and $B_0\bar{U}^TM\bar{V}$ are in $SPD_r$, we cannot directly prove that 
$\bar{B}=\textit{Sym}(\bar{U}^TM\bar{V})\in \textrm{SPD}_r$, but we can do it passing to 
the limit inside Algorithm \ref{Alg-FROpt}.
When passing to the limit the sequences defined by $\{U_i\}$, $\{B_i\}$ and $\{V_i\}$, both conditions are simultaneously 
met, as in Lemma~\ref{lemmaSPDLimit}:

\begin{lem}
Suppose that the \textit{FixedRankOptFull} Algorithm converges to a fixed point $(U^*,B^*,V^*)$, then
$U^*\in \textrm{St}_{m,r}$, $B^*\in \textrm{SPD}_r$, and $V^*\in \textrm{St}_{n,r}$.
\label{lemmaSPDLimit}
\end{lem}
%
\noindent \textit{Proof}: Since $\scalemath{0.92}{(U^*,B^*,V^*)=\textit{FixedRankStep}(M,U^*,B^*,V^*)}$, $U^*$ and $V^*$ are solutions to their respective OPPs and have to be in their respective Stiefel manifolds. Then, by applying Lemma~\ref{lemmaOPP}, both $U^*{}^TMV^*B^*$ and $B^*U^*{}^TMV^*$ are in SPD$_r$. Since $B^*=\textit{Sym}(U^*{}^TMV^*)$, we have that:
\begin{eqnarray}
 2B^*{}^2 &=& B^*\textit{Sym}(U^*{}^TMV^*)+\textit{Sym}(U^*{}^TMV^*)B^* \nonumber\\
&=& U^*{}^TMV^*B^*+B^*U^*{}^TMV^*\,,
\end{eqnarray}

\noindent which is on SPD$_r$ since it is a convex manifold. Then, by taking the square root of $B^*{}^2$ we have that 
$B^*\in \textrm{SPD}_r$. $\blacksquare$

Now, since the eigenvalues are continuous functions of the matrix entries, 
there exists $\epsilon>0$ such that all symmetric matrices in the open ball of radius $\epsilon$ centered at $B^*$  are contained in 
$\textrm{SPD}_r$. Thus, if \textit{FixedRankOptFull} converges, then there exists 
$n_0\in \mathbb{N}$ such that $B_i\in \textrm{SPD}_r\,\forall i\geq n_0$. 

%
%

Let us now discuss the convergence of the  \textit{FixedRankOptFull} Algorithm.
Given $S\in \textrm{St}_{p,k}$, then $P_S=SS^T$ is the projector onto the column space of 
$S$ in $\mathbb{R}^p$. Note that $P_S=P_{SQ}$, where $Q\in O_k$. Then we have the following:

\begin{thm}
 If $\text{rank}(MV_0)=r$, the \textit{FixedRankOptFull}  algorithm  converges Q-linearly to a global minimum of (\ref{eq:MinimizationFixedRank}) given by $(U^*,B^*,V^*)$ such that 
$L=U^*B^*V^*{}^T$ is the unique projection of $M$ onto ${\cal F}_{m,n}^{(r)}$. The convergence is Q-linear, in the sense that $||P_{U_{i}}-P_{U^*}||=\mathcal{O}((\frac{\sigma_{r+1}}{\sigma_r})^{2i})$ and 
$||P_{V_{i}}-P_{V^*}||=\mathcal{O}((\frac{\sigma_{r+1}}{\sigma_r})^{2i})$.
\label{TheoremGlobalMinimum}
\end{thm}

\noindent \textit{Proof}:
For each $U_i,V_i$, denote by $P_{U_i},P_{V_i}$ the projectors as defined before. Then it is easy to proof, using the alternative definition of $P_O[A]$,  that 
$P_{U_{i+1}}=P_{\tilde{U}_{i+1}}$, where $\tilde{U}_{i+1}=P_O[MM^TU_i]$. Thus the sequence of subspaces $\{P_{U_i}\}$ is the same as that produced by the Orthogonal Iteration~\cite{MatrixComputations} for the computation
of the first $r$ eigenvalues and eigenvectors of the symmetric matrix $MM^T$ . The Orthogonal Iteration converges Q-linearly in the sense that
$||P_{\tilde{U}_{i}}-P_{\tilde{U}^*}||=\mathcal{O}((\frac{\lambda_{r+1}}{\lambda_r})^i)$, with $\lambda_k$ the eigenvalues of $MM^T$. Since $\lambda_k=\sigma_k^2$, we have that 
$||P_{U_{i}}-P_{U^*}||=\mathcal{O}((\frac{\sigma_{r+1}}{\sigma_r})^{2i})$. By a similar argument  $||P_{V_{i}}-P_{V^*}||=\mathcal{O}((\frac{\sigma_{r+1}}{\sigma_r})^{2i})$. $\blacksquare$

In our case, $M=L+S$, with $L \in \mathcal{F}^{(r)}_{m,n}$ and $S$ is a perturbation matrix, then $\sigma_{r+1}$ will be much smaller than $\sigma_{r}$ and the error will be largely decreased in each iteration.


We would like to stress that although we do not provide an algebraic proof for $B_i \in \textrm{SPD}_r$ due to its complexity, 
Lemma \ref{lemmaSPDLimit} along with the continuity of eigenvalues argument guarantee that 
$B_i \in \textrm{SPD}_r$ when we are near an optimum. Starting with $B_0=I_r$ then for the first iteration $B_1$ 
we have $B_1=\textit{Sym}(U_1^TMV_1)= U_1^TMV_1\in \text{SPD}_r$,
and according to Figure \ref{fig:svd} very near the optimum, thus we can ensure that  the whole sequence $\{B_i\}$ is 
in $\textrm{SPD}_r$.
This is not a complete proof, but 
Theorem \ref{TheoremGlobalMinimum} ensures global convergence despite the nature of $B_i$. 
Thus at some point $B_i$ will be in $\textrm{SPD}_r$, which is also shown to always occur in our extensive numerical experiments, even starting from random $B_0$.


\balance
{\small
\bibliographystyle{ieee}
\bibliography{egbibShort}
}

\end{document}